\documentclass[runningheads]{llncs}
\usepackage[T1]{fontenc}
\usepackage{amsmath}
\usepackage{tabularx}
\usepackage{graphicx}
\begin{document}
\title{Real Time Captioning of Sign Language Gestures in Video Meetings}
\author{Sharanya Mukherjee\inst{1} \and
Md Hishaam Akhtar\inst{1} \and
Dr Kannadasan R\inst{1}}
\authorrunning{Sharanya Mukherjee, Md Hishaam Akhtar and Dr Kannadasan R}
\institute{Vellore Institute of Technology, Vellore}
\maketitle
\begin{abstract}
It has always been a rather tough task to communicate with someone possessing a hearing impairment. One of the most tested ways to establish such a communication is through the use of sign based languages. However, not many people are aware of the smaller intricacies involved with sign language. Sign language recognition using computer vision aims at eliminating the communication barrier between deaf-mute and ordinary people so that they can properly communicate with others. Recently the pandemic has left the whole world shaken up and has transformed the way we communicate. Video meetings have become essential for everyone, even people with a hearing disability. In recent studies, it has been found that people with hearing disabilities prefer to sign over typing during these video calls. In this paper, we are proposing a browser extension that will automatically translate sign language to subtitles for everyone else in the video call. The Large-scale dataset which contains more than 2000 Word-Level ASL videos \cite{12}, which were performed by over 100 signers will be used.
\keywords{machine learning \and deep learning \and image processing \and computer vision \and sign language \and browser extension \and socket}
\end{abstract}

\section{Introduction}
The use of sign based linguistics have become increasingly popular among humans with a hearing impairment. However, currently, there are only around 250,000 to 500,000 people using American Sign language which significantly limits the number of people with whom they can communicate with ease. The only substitute for sign language is through writing, which is quite cumbersome, impersonal, and impractical in real- life situations, especially during video meetings in this pandemic. To overcome this problem and to enable dynamic communication during video calls, we present a sign language recognition system that will automatically translate sign language to text in real-time and display it as subtitles for everyone in the video call.
\par
Sign language recognition where individual words will be identified at a time, and sign language recognition where individual characters will be identified at a time in real time. In this paper, we target word level American sign language identification in real- time using deep learning techniques. Our work will broadly consist of three tasks to be done in real-time:
\begin{itemize}
  \item \textbf{Video acquisition:} Obtaining the video of the person signing
  \item \textbf{Frame analysis and classification:} Analyzing and classifying the frames in the video
  \item \textbf{Prediction display:} Displaying the most likely predicted word
\end{itemize}
\par
From the perspective of computer vision, this work is significantly challenging due to various considerations including environmental concerns like lighting or camera position or background, occlusion like a part of hand being out of the range of vision of the camera, detecting the boundary between each signs (Detecting when a sign ends and the next begins)
\par
Previously neural networks were used to recognize character-level American sign language with accuracies consistently over 90 percent, but most of them require an element for 3-D capture through motion-tracking gloves or a similar costly hardware devices , and only very few of them provide real-time classifications. These restrictions imposed by hardware reduce the feasilbility and scalibility of these solutions.
\par
Our proposed system features a pipeline that takes video of the user who is signing through their local webcam. Next, individual frames are extracted from the video, and prediction is made on it by our model. The predicted word or character will be then broadcast over a socket connection into a room which the participants in the video meeting will be joining through our browser extension. The translated word or character is then displayed, as a subtitle for every user present in the meeting. This system enables dynamic real-time and hassle-free communication between hearing disabled people and others.
\par
To make a practical American Sign Language or ASL identification model, the training data needs to contain a massive amount of classes and training examples. For word- level American Sign Language recognition, we will be using the WLASL dataset \cite{12}, which contains a large-scale lexicon of signs and their corresponding annotations. The videos in the dataset are RGB-based videos that are scraped from the Internet, since we don’t want to anchorage the minimum hardware requirement for the sign identification. With this our trained sign language identification models will not have to depend on unique additional requirements such as Microsoft Kinect or special gloves.

\section{Related Work}
\par
Author Siming \cite{1} developed a new methodology to tackle the recognition of words. He constructed a dataset which contained 40 words which were frequently used in conversation. Also, he collected 10,000 related images. A region based convolutional neural network coupled with an embedded RPN or region proposal network module was used to detect the palms in action from the media. Such a technique bolstered performance of the model. Also, for the extraction of complex features, a three-dimensional CNN was used.
\par
Author Rekha J \cite{2} employed the use of YCbCr colour model to successfully recognize the human skin from the dataset. A principal curvature based region detector was employed for the extraction of minute features from the images. They were classified using standard algorithms like DTW, Multiclass SVM and a non linear KNN.
\par
In another paper \cite{3} , a color based technique was employed for inexpensive processing. The background of the images were set to green to enable the extraction of the skin from its green surroundings. The extracted results were then transformed into its greyscale counterparts. It was successful in mapping the hand gestures and produced a staggering accuracy of 92\%.
\par
M. Geetha along with U.C. Manjusha employed \cite{4} the use of individual alphabet and digits to recognize characters from the video. They extracted the skin region and removed its boundary to perform further analysis on it through a B-Spline Curve. Thereafter, a simple SVM was used for categorising the images. It produced a rather excellent accuracy of 90
\par
Author L. Pigou \cite{5} utilized the CLAP14 \cite{6} dataset which contained gestures from the Italian language. Pigou \cite{5} used a Convolutional Neural network after processing the images. The model had six layers for training and all the kernels are in 2D. The activation function which he used in his work is ReLU or Rectified Linear Units. CNN performs feature extraction. The classification was made possible through the employment of an ANN. The accuracy of the model was 91.70\% with an error rate of 8.3\%.
\par
Author J Huang \cite{7} built a dataset using Microsoft Kinect by using 25 frequently used words. A three-dimensional CNN was used with 5 main input channels. The aver- age accuracy was about 94.2\%.
\par
In a research work on gesture recognition done by author J.Carriera \cite{8}, the transfer learning method was used. ImageNet \cite{9} and Kinect \cite{6} datasets were used as the pre-trained datasets for his work. The RGB, flow and pre-trained kinetic was merged after training the model on multiple datasets. On the UCF-101 \cite{10} dataset, he achieved an accuracy of 98\% and 80.9\% on the HMDB-51 \cite{11} dataset.

\section{Implementation}
\subsection{Classifier Model}
\par
To achieve high precision and accuracy and low cost, transfer learning using MobileNetV2 was used.
\par
Transfer Learning is a unique approach where the models developed for a task are usually first trained on large data sets and then reconstructed to fit more specific data. This is usually achieved by reusing a portion of the pre-trained weights and then re initializing or altering weights at shallower layers. This technique is much faster computationally and needs fewer data requirements.
MobileNetV2 is an enhanced version of the MobileNetV1. Similar to most deep convolution networks, MobileNetV1 uses the activation function Rectified Linear Unit or popularly known as ReLU, which is described as:
\[
\text{ReLU}(x) = 
\begin{cases}
0 & \text{if } x < 0 \\
x & \text{if } x \geq 0
\end{cases}
\]
\par
Our classification is done using a convolution neural network or a CNN. Convolution Neural Network is a machine learning algorithm that is very important for tasks related to image or video processing. A very big advantage of using this technique is CNN’s ability to produce high accuracies. CNN is a hierarchical model where different layers of neurons are connected together and similar to other machine learning algorithms, it tries to optimize the loss function. In our work, we have used a softmax-based loss function.

\begin{figure}
\includegraphics[width=\textwidth]{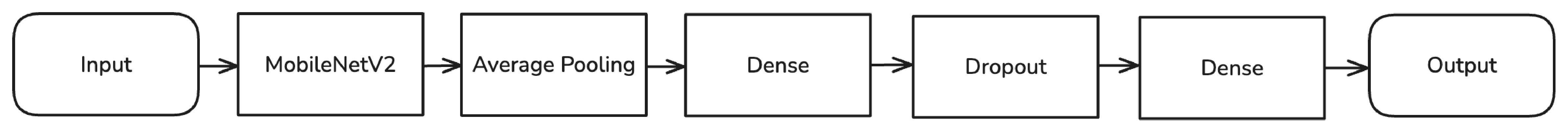}
\caption{Proposed Model Architecture} \label{fig1}
\end{figure}

\par
The weights which were trained on an ImageNet was first fed to the MobileNetV2. One thousand neurons were used in a neural network layer which was connected to the MobileNetV2.
\par
The MobileNetV2 was first pre-loaded with weight trained on ImageNet. We added an average pooling layer to the MobileNetV2 output feature map, with a dense neural network layer containing 1000 neurons. We used the standard ReLU as our activation function. We added a 75\% dropout layer to prevent over fitting. Our final layer is the dense layer that classifies the given frames and it had a softmax activation function given by:
\[
\text{Softmax}(z_i) = \frac{e^{z_i}}{\sum_{j=1}^{K} e^{z_j}} \quad \text{for } i = 1, 2, \dots, K
\]
\subsection{Experimenting different models}
\par
Initially we used transfer learning techniques with Inception 3D or I3D algorithm, which is used for video classification in three dimension, to classify our sign language videos. The I3D model is trained on ImageNet after which it has been fine-tuned on Kinetics-400. The original one has 400 classes and three input channels. Now in our dataset we have 2000 classes, so we modified the original I3D model such that the last layer has 2000 neurons instead of 400 as the Kinetics dataset it was initially trained on had 400 human actions collected from YouTube videos.
\par
In this approach, we first converted the videos in our dataset into mp4 format after which we extracted individual frames from the video instance to train upon. We used around eighteen thousand videos to train upon and approximately three thousand videos to test our trained model upon. With this approach we got a total accuracy of 40.5\%.
\par
In the next approach, we first performed principal component analysis followed by using a support vector machine classifier. Principal component analysis or PCA is used for dimension reduction of a large dataset, where a big set of features is turned into a smaller one which still contains all the important information of the larger dataset. In our work, we used Principal component analysis to extract features which we implement using the scikit-learn library. These reduced number of features were then used to train our SVM model. Support vector machine is a type of supervised machine learning algorithm which is primarily used for classification tasks. Here in an n-dimensional space, each data item is plotted as points, and then we find a hyper plane which divides the classes in an efficient way.
\par
With this approach we got an accuracy of this approach we got an accuracy of 59.8\% when trained on 100 classes of data.
\subsection{Accuracies Recorded}
Our models showed promising results when tested on our data and below are the accuracies recorded for each algorithm:

\begin{table}[htbp]
\caption{Table showing accuracies recorded for each algorithm}
\label{tab1}
\begin{tabularx}{\textwidth}{|X|c|c|}
\hline
\bfseries Classification Approach & \bfseries Number of classes & \bfseries Maximum Accuracy\\
\hline
Transfer Learning Using MobileNetV2 & 2000 & 63\%\\
Transfer Learning Using Inception-3D Model & 2000 & 40.5\%\\
Support Vector Machine along with PCA & 100 & 59.8\%\\
\hline
\end{tabularx}
\end{table}

\subsection{Pipeline}
\par
Before training our model, we need to process the video. For training, we will be splitting each video data into 12 frames. But some videos are shorter in size, hence we made a custom video extender for them, with which those videos will also be able to generate a consistent number of frames.
\par
After the implementation of the model, the next we developed was an extension for Google Chrome so that the model could be deployed and used for video calls. To overcome the challenge of dealing with multiple platforms, we constructed a unique architecture for the chrome extension. It consists of three components: the chrome extension, a message broadcasting server and a local desktop application. Whenever a specially abled person has to start speaking, they turn on the desktop application in the background. The desktop application picks up the frames of the image and deciphers the words using the machine learning model. Once the words or letters are extracted, they are sent to the broadcasting server using a real time socket connection. In a socket connection two or more nodes are connected in a network where information exchange takes place in real time. This chrome extension connects to the broadcasting server. The broadcasting server, after receiving the words from the desktop application broadcasts it to the chrome extension which shows up in the screen of everyone in the current meeting. This is how we were able to process the images very fast and show it to everyone in the meeting in real time. Here is a diagram of the architecture:

\begin{figure}
\includegraphics[width=\textwidth]{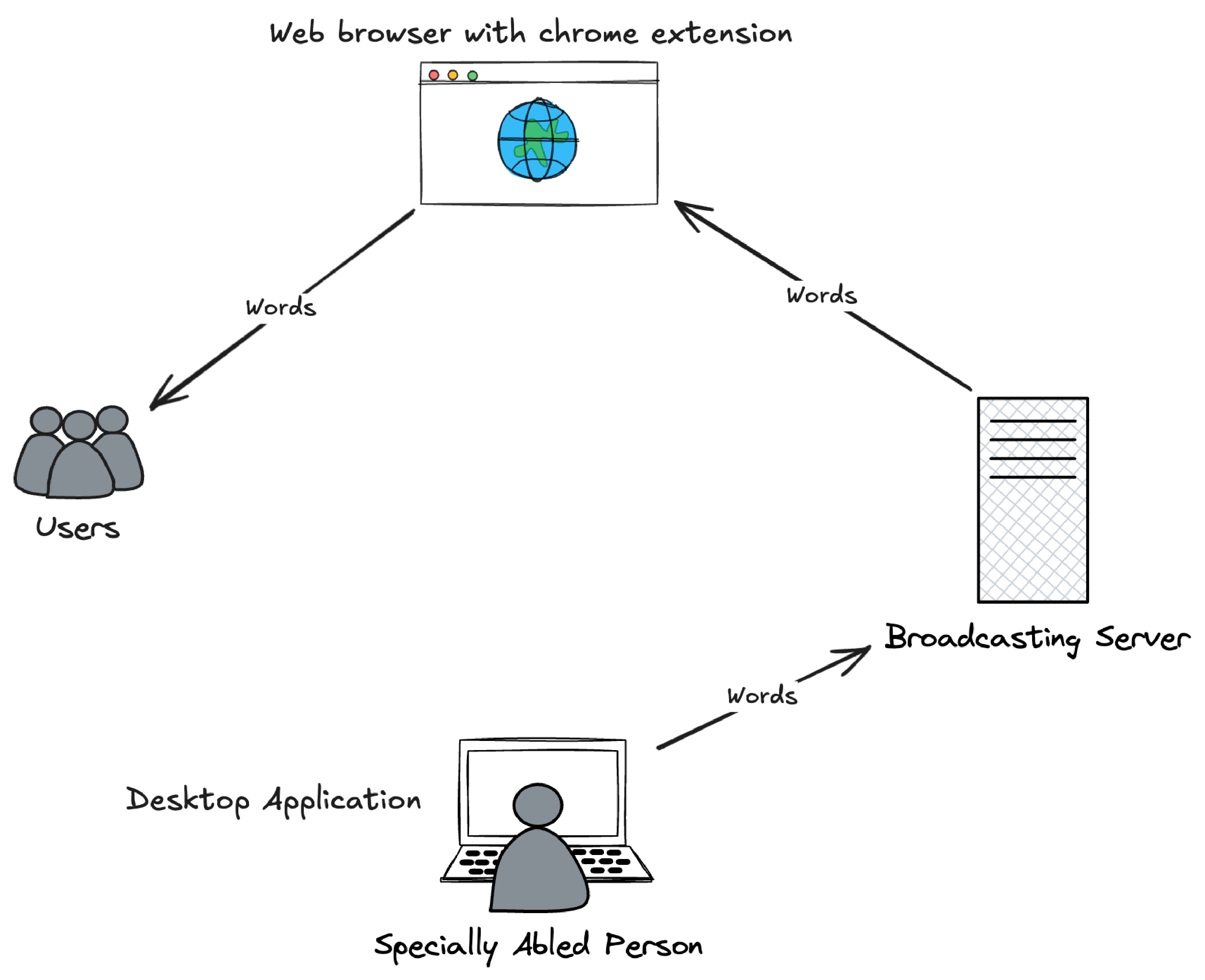}
\caption{Proposed System Architecture} \label{fig2}
\end{figure}

\subsubsection{Acknowledgements} 
We wish to acknowledge Vellore Institute of Technology and Dr. Kannadasan R for guiding us through our research.

\end{document}